\documentclass[a4paper, 10pt, conference]{ieeeconf}      

\IEEEoverridecommandlockouts                              

\usepackage{graphicx} 
\usepackage{amsmath} 
\usepackage{amssymb}  
\usepackage{multicol}  
\usepackage{multirow}  

\title{\LARGE \bf
Attention-based Supply-Demand Prediction for Autonomous Vehicles*
}

\author{Zikai Zhang$^{1,2}$, Yidong Li$^{2}$, Hairong Dong$^{1}$, Yizhe You$^{2}$ and Fengping Zhao$^{2}$
\thanks{*This work was supported by the National Natural Science Foundation of China under Grant No. 61672088 and No. 61790573}
\thanks{$^{1}$School of Electronic and Information Engineering, Beijing Jiaotong University, Beijing 100044, China
        {\tt\small hrdong@bjtu.edu.cn}}%
\thanks{$^{2}$School of Computer and Information Technology, Beijing Jiaotong University, Beijing 100044, China
        {\tt\small ydli@bjtu.edu.cn}}%
}

\begin{document}

\maketitle
\thispagestyle{empty}
\pagestyle{empty}

\begin{abstract}

As one of the important functions of the intelligent transportation system (ITS), supply-demand prediction for autonomous vehicles provides a decision basis 
for its control. In this paper, we present two prediction models (i.e. ARLP model and Advanced ARLP model)
based on two system environments that only the current day's historical data is available or several days' historical data are available.  These two models jointly consider the spatial, temporal, and semantic relations. Spatial dependency is captured with residual network and dimension reduction. Short term temporal dependency is captured with LSTM. Long term temporal dependency and temporal shifting are captured with LSTM and attention mechanism. Semantic dependency is captured with multi-attention mechanism and autocorrelation coefficient method. Extensive experiments show that our frameworks provide more accurate and stable prediction results than the existing methods.
\end{abstract}

\section{INTRODUCTION}

Due to the rapid growth of the urban population, the road is more crowd with vehicles, and thus result in environmental problem, traffic safety problem and traffic congestion problem. In order to solve these problems, the Autonomous Vehicle (AV) is gaining increasing attention from the public and research community. Besides autonomous driving capabilities, inter-vehicle communications \cite{1 J} can further enhance the AVs with shared information. Moreover, the control center can be used to systematically coordinates and schedules AVs with an optional ride-sharing capability to improve social welfare \cite{2 J}, to reduce traffic congestion and to enhance transportation throughput.    
These exist many pricing mechanisms based control center\cite{3 A}\cite{4 J}\cite{5 Yu}, both of them need the information about the gap between AV supply and demand. Therefore, an accurate supply-demand prediction method is needed for control centers to percept and predict the gap between AV supply and demand.

Supply-demand prediction is a kind of traffic flow prediction, which aims to predict the traffic for the next time slot when given historical traffic data (e.g., traffic volume, traffic speed, traffic demand).
Recently, deep learning has shown its great learning and representation power in traffic flow prediction. To capture spatial correlation, researchers propose to use convolutional structure.
For example, Zhang et al. \cite{6 Zhang} constructed three same convolutional structures to capture trend, period and closeness information. Ma et al. \cite{7 Ma} utilized CNN on the whole city for traffic speed prediction. Yao et al. \cite{8 Yao} proposed a flow-gated local CNN to handle spatial dependency by modeling the dynamic similarity among locations using traffic flow information. 
To model non-linear temporal dependency, researchers propose to use recurrent neural network based framework. For example, Yu et al. \cite{9 Yu} applied Long-short-term memory (LSTM) network to capture the sequential dependency for predicting the traffic under extreme conditions. Li et al. \cite{10 Li} utilized graph convolutional GRU for traffic speed prediction. To capture both longterm periodic information and temporal shifting, researchers propose to use weighted representation. For example, Yao et al. \cite{8 Yao} proposed a periodically shifted attention mechanism by taking long-term periodic information and temporal shifting simultaneously. Wang et al. \cite{11 deepsd} used a Softmax Layer to get the weight vector of representation.

Only a few works attempt to deal with Semantic similarity to enhance the prediction accuracy. Yuan et al. \cite{12 clusting-semanic} treated the clustering stage as a separate sub-task and manually designed the distance measure, which is a non-trivial task. Wang et al. \cite{11 deepsd} and Yao et al. \cite{13 dmvst-net} proposed the end-to-end network to optimize the embedding parameters together with other parameters. The parameters are optimized through backpropagation towards minimizing the final prediction loss. But, the semantic similarity learned by embedding method contains some irrelevant information, which will impact the improvement of prediction performance.
Moreover, the pseudo-periodic traffic signals have random white noise.  Dunne et al. \cite{14 autocorr} solve this problem with Autocorrelation coefficient method in a neuro wavelet model. In this paper,  we propose a novel network to automatically get semantic similarity with weighting parameters and use multi-attention-based network component to further reduce the white noise.


In this paper, we propose two prediction models for Autonomous Vehicles with different system environment. For short temporal historical data, ARLP model is proposed with LSTM to capture temporal dependency. For long temporal historical data, Advanced ARLP model is proposed with LSTM and attention mechanism to capture temporal dependency and temporal shifting. 

Other contributions are summarized as follows:
\begin{itemize}
	
	\item We proposed two models based on the system environment that jointly considers the spatial, temporal, and semantic relations. 
	\item We proposed a semantic component that captures semantic dependency and reduces the impact of random white noise by multi-attention mechanism and autocorrelation coefficient method. In addition, the semantic component automatically learns the similarity distance across four aspects.
	\item We proposed a spatial component that captures spatial dependency with dimension reduction.
	\item We conducted extensive experiments on a large-scale online car-hailing dataset. The results show that our methods outperform the competing baselines in both short temporal historical data and long temporal historical data.
	
\end{itemize}

The rest of the paper is organized as follows. We describe some notations and our problem
in Section II. The proposed methods are presented in Section III. We describe the experimental results in detail in Section IV. Finally, we conclude the paper in Section V.

\section{PRELIMINARIES}

In this section, we first fix some notations and define the demand-supply prediction problem. We follow previous studies \cite{8 Yao}\cite{13 dmvst-net} and split the whole city to an $X\times Y$ grid map which consists
of $X$ rows and $Y$ columns. define the set of non-overlapping locations $L = \left\{ l_{1}, l_{2}, ..., l_{i}, ..., l_{X \times Y} \right\}$. Also, we split time as $T$ nonoverlapping time intervals and define the time interval set as $T = \left\{ t_{1}, t_{2}, ..., t_{j}, ..., t_{T} \right\}$. We define the current day is $d_{D}$, and the history day set is defined as $D = \left\{ d_{1}, d_{2}, ..., d_{k}, ..., d_{D} \right\}$. 

\textbf{Weather Information:} For a specific area $l_{i}$ at timeslot $t_{j}$, the weather condition is defined as $Wea_{j}^{i}$ (weather type: rainy, sunny, cloudy, etc.).

\textbf{Traffic Speed:} For a specific area $l_{i}$ at timeslot $t_{j}$, the traffic speed is the average AV speed and defined as $Ts_{j}^{i}$.

\textbf{Traffic Volume:} For a specific area $l_{i}$ at timeslot $t_{j}$, the traffic Volume is the average amount number of AV and defined as $Tv_{j}^{i}$.

\textbf{Journal Distance (u):} For a specific area $l_{i}$ at timeslot $t_{j}$, the journal distance (pick up) is the average distance of riders who start a journal in area $l_{i}$ and defined as $Ju_{j}^{i}$. 

\textbf{Journal Distance (d):} For a specific area $l_{i}$ at timeslot $t_{j}$, the journal distance (drop off) is the average distance of riders who end a journal area $l_{i}$ and defined as $Jd_{j}^{i}$.

\textbf{Demand Volume:} For a specific area $l_{i}$ at timeslot $t_{j}$, the demand volume is the amount number of e-hailing request for AVs and defined as $D_{j}^{i}$.

\textbf{Supply Volume:} For a specific area $l_{i}$ at timeslot $t_{j}$, the supply volume is the amount number of AVs which have capacity to provide service and defined as $S_{j}^{i}$.

\textbf{Demand-supply Gap:} For a specific area $l_{i}$ at timeslot $t_{j}$, the
demand-supply gap is demand volume subtract supply volume (i.e. $D_{j}^{i} - S_{j}^{i}$) and defined as $DS_{j}^{i}$.

\textbf{Prediction Problem:} Suppose the current date is  $d_{D}$ day and the current time slot is $t_{T}$. Given the historical data, our goal is to predict the supply-demand
gap $DS_{T}^{i}$ for target area $l_{i}$.

In this paper, we consider two situations to prediction based on the system environment. The first situation is that only the current day's historical data is available.  To solve this problem, the attention-based supply-demand prediction model (ARLP) is proposed. The second situation is that $D$ days' historical data are available. To solve this problem, an advanced version of attention-based supply-demand prediction model (Advanced ARLP) is proposed.


\section{METHODS}


In this section, we provide details for our proposed models. Our model is divided into three parts, i.e., semantic representation component,  spatial representation component, temporal representation and prediction component. Using semantic representation component, we can find the semantic dependency by discovering the similarities among different areas and reducing the random white noise. Using spatial representation component, we can find spatial dependency by convolution and dimensionality reduction. Using temporal representation and prediction component, we can find the temporal dependency and predict the result based on different kind of historical data.

\subsection{Semantic Component}
\begin{figure}
	\centering
	\includegraphics[width=0.9\linewidth]{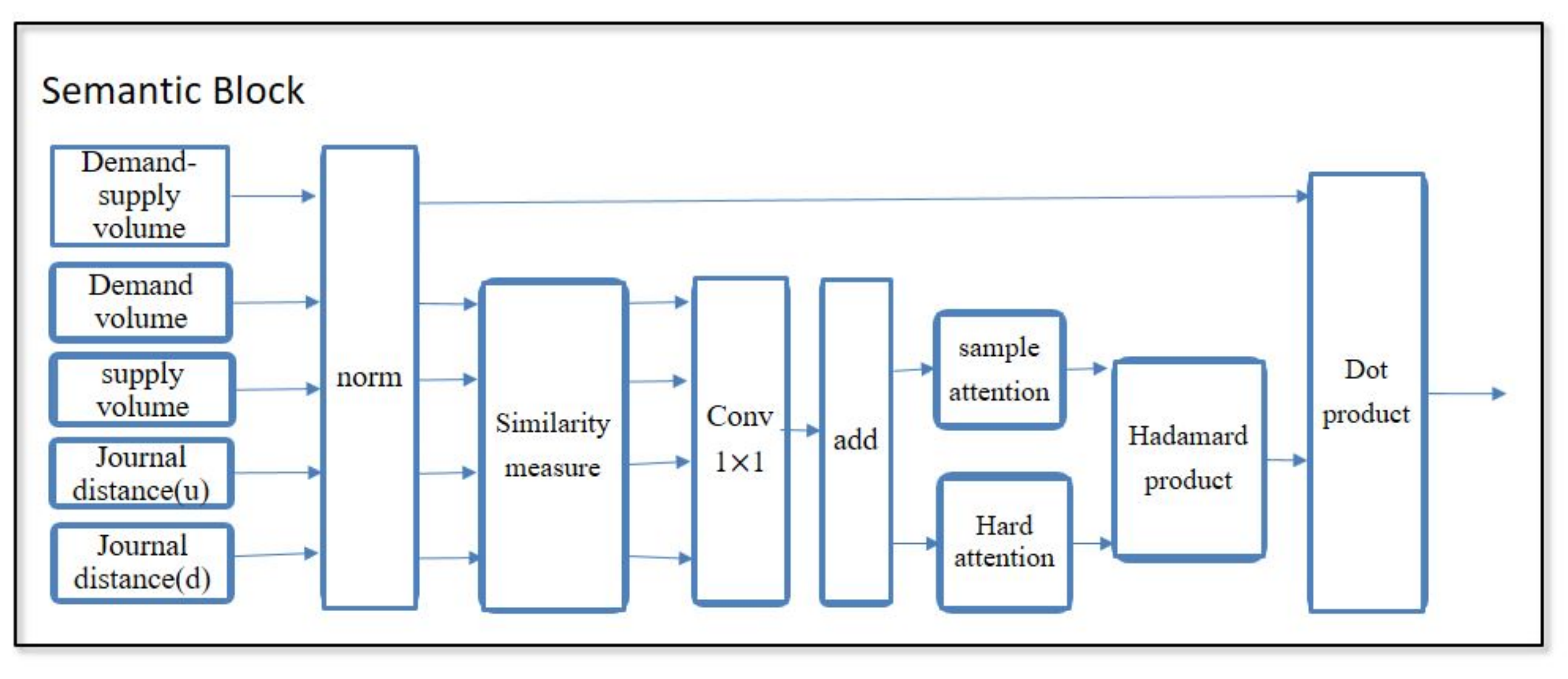}
	\caption{Semantic Representation Component}
	\label{networkfigure1}
\end{figure}

In order to automatically learn similarity distance, capture the semantic dependency and reduce the impact of random white noise (such as traffic accident), we propose the semantic representation component, as shown in Figure \ref{networkfigure1}.

As shown in Figure. \ref{networkfigure1}, we treat the whole city as five $X \times Y$ images. These five images contain demand-supply gap information, demand information, supply information, Journal Distance (u) information, and Journal Distance (d) information respectively.  After normalization, we can get matrices $M_{ds}$, $M_{dv}$, $M_{sv}$, $M_{ju}$ and $M_{jd}$ respectively. Every element in matrix $M$ is a time-series vector which contains information within the time series $\left\{ t_{1}, t_{2}, ..., t_{j}, ..., t_{T} \right\}$.

In order to capture the most similar trend representation with reduced impact of the random white noise, the autocorrelation coefficient vector \cite{14 autocorr} is used to convert data representation. 
In the similarity measure stage, we use four aspects of matrices ($M_{dv}$, $M_{sv}$, $M_{ju}$, and $M_{jd}$) to measure the similarity distance. Calculating the autocorrelation coefficient vector, we can get four matrices, i.e., $M_{dv}^{ac}$, $M_{sv}^{ac}$, $M_{ju}^{ac}$ and $M_{jd}^{ac}$.

For a specific area $l_{i}$ at timeslot $t_{j}$, the similarity is measured as $s_{ij} = m_{ij}^{ac} \cdot m_{kj}^{ac}$, where $l_{k}$ is the target prediction location. Finally, we get four similarity matrices ($S^{dv}$, $S^{sv}$, $S^{ju}$, $S^{jd}$).

In order to measure the similarity distance with these four matrices, we use $1 \times 1$ convolution layer to automatically assign weights. After adding these four weighted matrices together, we can get the similarity distance matrix $SD = \left\{ sd_{1}, sd_{2}, ..., sd_{i}, ..., sd_{X \times Y} \right\}$.

In order to capture the periodic traffic wavelet in the semantic aspect, we propose a hard attention  mechanism to filter out the semantic locations. 
In hard attention layer, the hard attention is calculated as follows,
$$ha= \begin{cases}
1 & \text{, if } sd_{i} > \beta sd_{k}\\ 
0 & \text{, otherwise } 
\end{cases}$$
where $\beta$ is a weight to get threshold value, $l_{k}$ is the target location and $l_{i}$ is the current location. In this paper, $\beta$ is set as 0.9.

In order to get the prediction model with stationarity, the semantic similar wavelets should be synthesized to one. Then, we propose a sample attention mechanism to assign weights for every wavelet.
The sample attention is calculated as follows,  $$sa_{i}= sd_{i}/sd_{k}$$ where  $l_{k}$ is target location and $l_{i}$ is current location.

Using Hadamard product for the sample attention and the hard attention, we can get the final attention $fa=ha*sa$. Using dot product with $M_{ds}$, we get the synthesized semantic representation of demand-supply gap, denoted as $S=\left\{ s_{j}^{i} \right\}$.

\subsection{Spatial Component}
\begin{figure}
	\centering
	\includegraphics[width=0.9\linewidth]{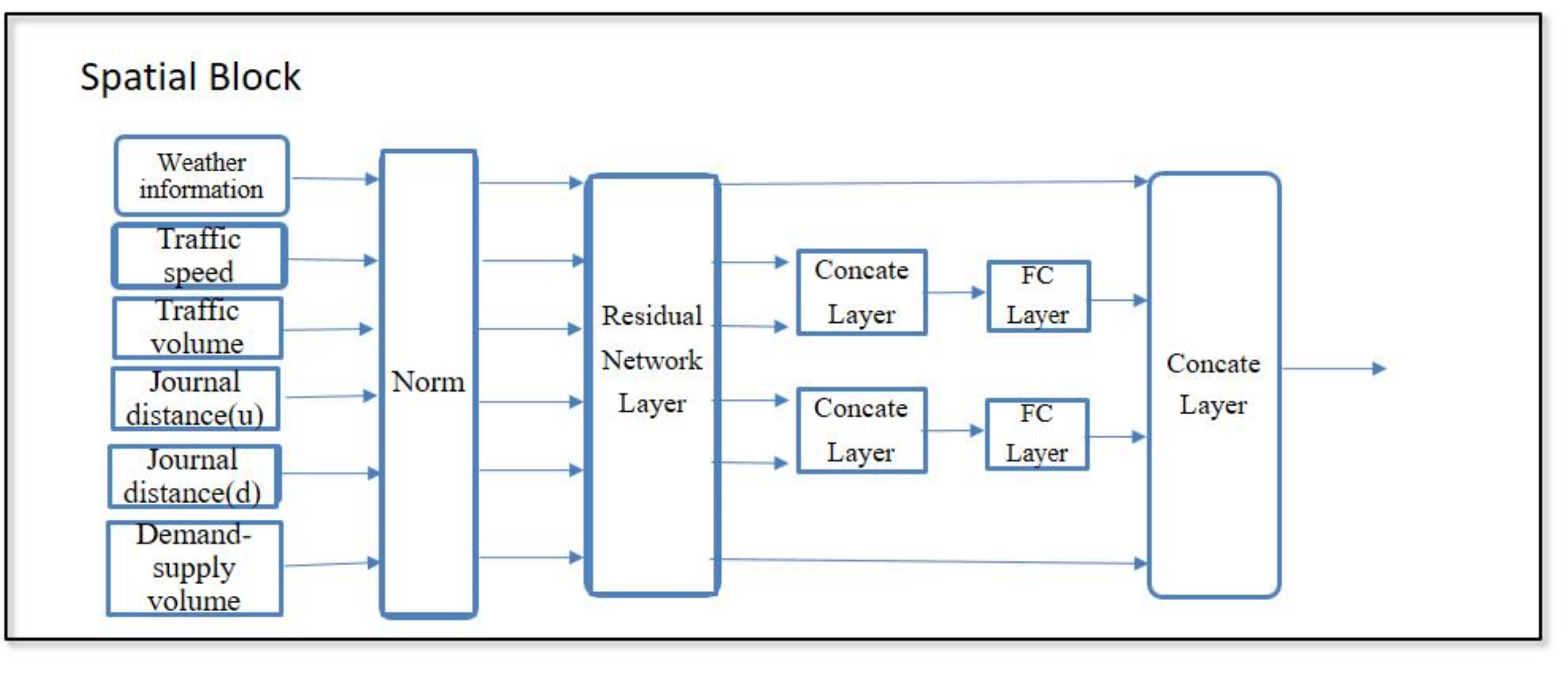}
	\caption{Spatial Representation Component}
	\label{networkfigure2}
\end{figure}

In order to capture the spatial dependency, we propose the spatial representation component, as shown in Figure. \ref{networkfigure2}.

Following previous studies \cite{8 Yao}\cite{13 dmvst-net}, we treat the target region $l_{i}$ and its surrounding neighbors as an $S \times S$ image for each time interval $t_{j}$. Then, we use the residual network to capture spatial representations respectively (denote as, $Gwea_{i}^{i}$, $Gts_{j}^{i}$, $Gtv_{j}^{i}$, $Gju_{j}^{i}$, $Gjd_{j}^{i}$ and $Gds_{j}^{i}$) after normalization.

In order to learn a better representation, dimensionality reduction is applied. We concatenate two pairs of high spatial dependency representations ($Gts_{j}^{i}$ and $Gtv_{j}^{i}$, $Gju_{j}^{i}$ and $Gjd_{j}^{i}$) together and use FC layer to reduce dimension, respectively. 

Finally, we concatenate these spatial representations together to get the final spatial representation $g_{j}^{i}$.

\subsection{Temporal Representation and Prediction Component}
\begin{figure}
	\centering
	\includegraphics[width=0.9\linewidth]{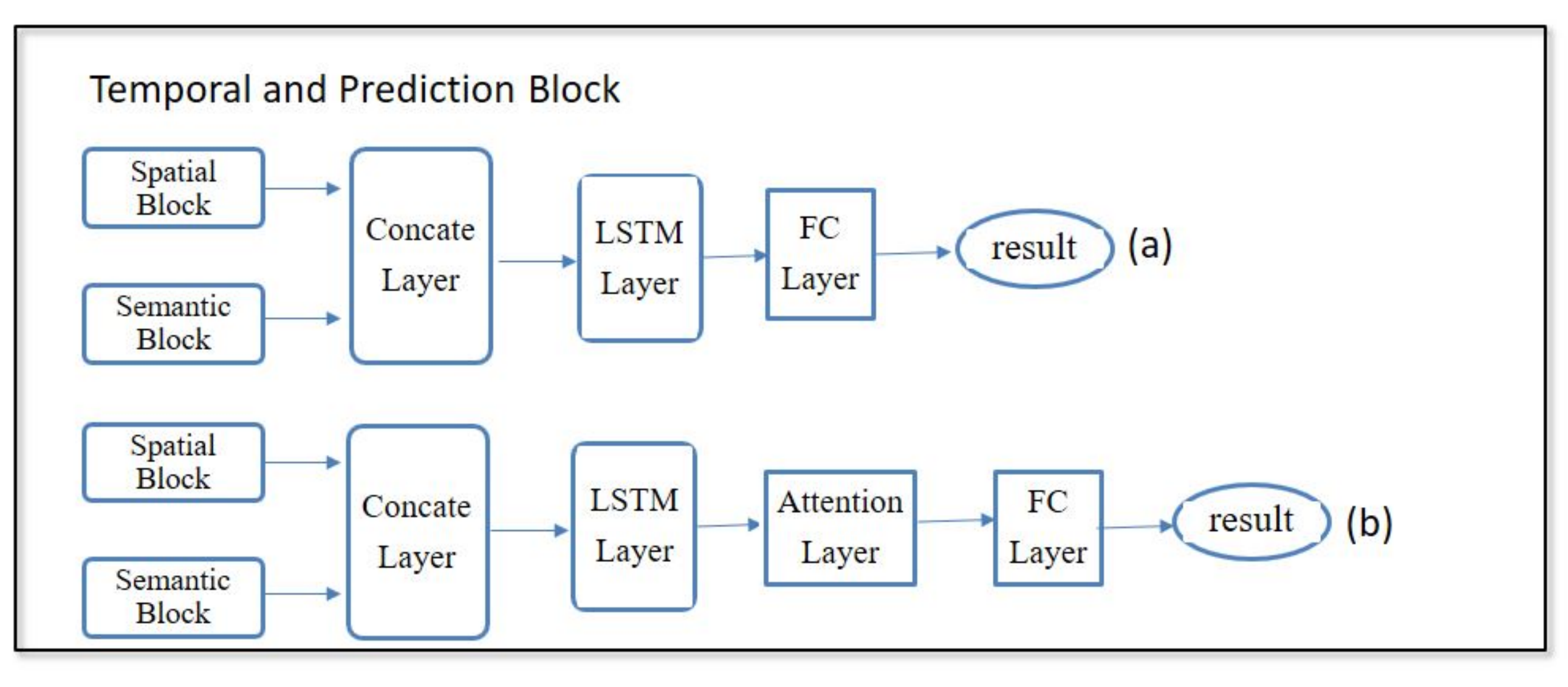}
	\caption{Temporal Representation and Prediction Component}
	\label{networkfigure3}
\end{figure}

In this subsection, we consider two situations to prediction based on the system environment. When only the current day's historical data is available, the Attention Based Supply-Demand Prediction model is proposed, as shown in Figure. \ref{networkfigure3} (a). When $D$ days' historical data are available, an advanced version of Attention Based Supply-Demand Prediction model is proposed, as shown in Figure. \ref{networkfigure3} (b).

\subsubsection{\textbf{Prediction model with the current day's data}} When concatenating the output of the spatial block and the semantic block ($S=\left\{ s_{j}^{i} \right\}$ and $G1=\left\{ g_{j}^{i} \right\}$) together, we can get the current day's time series data $F$.

Using LSTM, we can get the output of the last time interval $T$ as the representation
of temporal dependency, denoted as $h_{T}^{i}$. 

Then we feed $h_{T}^{i}$ to the fully connected network to get the final prediction value $\hat{y}_{T+1}^{i} $ for region $l_{i}$. The final prediction function is,
$$\hat{y}_{T+1}^{i} = RELU(W_{f}h_{T}^{i}+b_{f}) $$
where $W_{f}$ and $b_{f}$ are learnable parameters, $RELU(\cdot)$ is Linear activation function.

\subsubsection{\textbf{Prediction model with several day's data}} For the $k$-th day, when concatenating the output of the spatial block and the semanic block ($S=\left\{ s_{j}^{i} \right\}$ and $G1=\left\{ g_{j}^{i} \right\}$) together, we can get the $k$-th day's time series data $F_{k}$.
For the historical day set, the time series data set is $\mathbb{F}= \left\{ F_{k} \right\}$.

 Using LSTM, we can get the output of each day's last time interval $T$ as the representation
 of temporal dependency, denoted as $H_{T}^{i}= \left\{ h_{T,k}^{i} \right\}$.

Similar with \cite{7 Ma}\cite{13 dmvst-net}, we can use the attention mechanism to overcome temporal shifting.
In this paper, we can get the content-based attention, which is defined as, $$\alpha_{T,k}^{i} = \frac{h_{T,k}^{i} \cdot h_{T,D}^{i}}{\sum_{m=1}^{D}h_{T,m}^{i} \cdot h_{T,D}^{i}}.$$
With $\alpha_{T,k}^{i}$, we can get the long term representation, which is defined as,
$$h_{T}^{i,L} = \sum_{k=1}^{D} \alpha_{T,k}^{i} \cdot h_{T,k}^{i}.$$ 

We joint the long term representation and the short term representation ($h_{T,D}^{i}$ and $h_{T}^{i,L}$) together using concatenate layer, and defined as $h_{T}^{i}$.
Then we feed $h_{T}^{i}$ to the fully connected network to get the final prediction value $\hat{y}_{T+1}^{i} $ for region $l_{i}$. The final prediction function is,
$$\hat{y}_{T+1}^{i} = RELU(W_{f}h_{T}^{i}+b_{f}) $$
where $W_{f}$ and $b_{f}$ are learnable parameters, $RELU(\cdot)$ is Linear activation function.

\subsubsection{\textbf{Loss Function}}The loss function we used is defined as:
$${L}(\theta) = \sum_{i=1}^{N} (\hat{y}_{T+1}^{i} - y_{T+1}^{i})^2 $$
where $N$ is the batch size.

\section{EXPERIMENTS AND ANALYSIS}
In this section, we conduct experiments on the real dataset. We show a comprehensive quantitative evaluation by comparing with other methods and also show the effectiveness of the semantic representation learning.

\subsection{Dataset and Parameters} 

We evaluate our proposed methods on the dataset collected from the online car-hailing company in China. The dataset contains taxi order data, trajectory data and weather data for the city of Beijing. The taxi order data contains request start location, request end location, request start time, request end time, et. al. The taxi trajectory data contains location, speed, direction for every 2 seconds. The weather data contains weather information such as rainy, sunny, cloudy for every 30 minutes.

We split the whole city as 20 $\times$ 10 regions. The length of each time interval is set as 30 minutes. We statistic traffic speed, traffic volume, journal distance, demand, supply information for every 30 minutes. 

For training and testing, we divide the data into a 5:3 training set and a test set according to the time interval. When testing the prediction result, we use the previous 2.5 hours to predict the result for the next half hour.  For the situation that many days' historical data are available to predict, we set the available time is 5 days.

We set all convolution kernel sizes to 3 $\times$ 3. The size of each neighborhood considered is set as 5 $\times$ 5. The batch size in our experiment is set to 64 and the learning rate is set as 0.001. The residual network is set as 4 layers and the fully connected network is set as 3 layers.

\subsection{Evaluation Metrics}
In our experiment, we use Mean Average Percentage Error (MAPE), Mean Absolute Error (MAE) and Rooted Mean Square Error (RMSE) as the evaluation metrics. These metrics are defined as follows,
$$MAE = \frac{1}{N} \sum_{i = 1}^{N} \left | \hat{y}_{T+1}^{i} - y_{T+1}^{i}  \right | $$
$$RMSE = \sqrt{\frac{1}{N} \sum_{i = 1}^{N} (\hat{y}_{T+1}^{i} - y_{T+1}^{i})^2  }$$
$$MAPE = \frac{1}{N} \sum_{i = 1}^{N}  \frac{\left | \hat{y}_{T+1}^{i} - y_{T+1}^{i}  \right |}{\left |y_{T+1}^{i}  \right |} $$
where $\hat{y}_{T+1}^{i}$ and $y_{T+1}^{i}$represent the prediction value and the real value of region $l_{i}$ for time interval $T + 1$, and $N$ is the total number of samples.

\subsection{Compared Algorithms}
ARIMA\cite{17 arima}: ARIMA considers moving average and autoregressive components with the current day's demand-supply data.

LSTM: only short term temporal with the current day's demand-supply data is considered.

ConvLSTM\cite{15 convlstm} : ConvLSTM extends the fully connected LSTM to have convolutional structures in each transition with the current day's data. 

Reduced-ConvLSTM: Reduced-ConvLSTM is ConvLSTM model with spatial representation component. In this paper, ConvLSTM model can be treated as ARLP model without semantic representation component.

ST-ResNet\cite{18 ST-resnet}: ST-ResNet is a CNN-based deep learning framework for traffic prediction with many days' data.

DMVST-Net\cite{13 dmvst-net}: DMVST-Net is a deep learning based model which consists of three views:
temporal view (use LSTM), spatial view (use local CNN), and semantic view (use semantic network embedding). There are three part of differences between advanced ARLP and DMVST-Net: temporal shifting, dimension reducing and semantic similarity based attention. 

STDN\cite{16 stdn}: STDN is a deep learning based model which considers temporal shifting. In this paper, STDN is applied without flow gating.

Reduced-STDN: Reduced-STDN is STDN model with spatial representation component.

\subsection{Performance}

Table I and II show the performance of our proposed methods as
compared to all other competing methods, respectively. We run each baseline 10 times and
report the average results of each baseline. Our proposed methods
outperform all competing baselines by achieving MAE, RMSE, and
MAPE.

For the situation that only the current day's historical data is available, the Attention Based Supply-Demand Prediction model (ARLP) performs better than baselines,  as shown in Table I. The comparison between ARLP and Reduced-ConvLSTM proves the efficiency of our proposed semantic representation component for short temporal dependency. The comparison between Reduced-ConvLSTM and ConvLSTM proves the efficiency of our proposed dimension reducing method in spatial representation component for short temporal dependency. But our dimension reducing method is more sensitive to abnormal data.

\begin{table}[h]
	\caption{Comparison with Different Baselines: short temporal}
	\label{table_1}
	\begin{center}
		\begin{tabular}{llll}
			\hline
			\multirow{2}{*}{MODEL} & \multicolumn{3}{c}{Metrics} \\ \cline{2-4}
		        	&  MAE      & RMSE       &  MAPE       \\ \hline
			ARIMA	&  0.2483   &   0.4334   &   41.81\%      \\ \hline
			LSTM	&  0.1680   &  0.2104    &  35.56\%       \\ \hline
		CONVLSTM	&  0.1565   &  	0.2073   &  31.77\%       \\ \hline
 Reduced-ConvLSTM   &  0.1512   &  	0.2101   &  30.71\%       \\ \hline
		    ARLP	&  0.1361   &  	0.1485   &   30.67\%      \\ \hline
		\end{tabular}
	\end{center}
\end{table}

For the situation that several days' historical data are available, the advanced version of Attention Based Supply-Demand Prediction model (Advanced ARLP) performs better than baselines,  as shown in Table II. The comparison between Advanced ARLP and Reduced-STDN proves the efficiency of our proposed semantic similarity based attention mechanism in semantic representation component for long temporal dependency. The comparison between Reduced-STDN and STDN proves the efficiency of our proposed dimension reducing method in spatial representation component for long temporal dependency.
It also can be seen in this situation that our dimension reducing method is more sensitive to abnormal data. 
\begin{table}[h]
	\caption{Comparison with Different Baselines: long temporal}
	\label{table_2}
	\begin{center}
		\begin{tabular}{llll}
			\hline
			\multirow{2}{*}{MODEL} & \multicolumn{3}{c}{Metrics} \\ \cline{2-4}
		            	&  MAE      & RMSE       &  MAPE       \\ \hline
			ST-ResNet	&  0.1489   &  0.2242    &   23.79\%      \\ \hline
			DMVST-Net	&  0.1438   & 0.1774     &  21.76\%       \\ \hline
		    	STDN	&  0.1348   &  	0.1634   &  21.08\%       \\ \hline
	    	Reduced-STDN&  0.1129   &  	0.1652   &  20.36\%       \\ \hline
		Advanced ARLP	&  0.1086   &  	0.1225   &   15.48\%      \\ \hline
		\end{tabular}
	\end{center}
\end{table}

For our proposed spatial representation component, we can get more accurate spatial representation, despite the weakness of sensitive with abnormal data. When together with our semantic representation component, we can get a good performance in accuracy and stability.


\section{CONCLUSIONS}
In this paper, we propose two models for supply-demand prediction based on the system environment. ARLP is designed for short temporal historical data. Advanced ARLP is designed for long temporal historical data. Our methods capture the spatial representation with a CNN-based network and dimension reduction method. Our methods capture semantic representation with multi-attention mechanism and autocorrelation coefficient method. ARLP captures short temporal dependency with LSTM. Advanced ARLP captures long temporal dependency with attention mechanism and LSTM. 
The evaluations show that our proposed models significantly outperforms the state-of-the-art
methods.

\addtolength{\textheight}{-12cm}   


\end{document}